\documentclass[10pt,twocolumn,letterpaper]{article}

\usepackage{cvpr}              

\usepackage[accsupp]{axessibility}  

\usepackage{graphicx}
\usepackage{amsmath}
\usepackage{amssymb}
\usepackage{booktabs}
\usepackage[dvipsnames]{xcolor}
\usepackage{makecell}
\usepackage{siunitx}
\usepackage{multirow}
\usepackage[flushleft]{threeparttable}
\usepackage{caption}

\usepackage[pagebackref,breaklinks,colorlinks]{hyperref}

\usepackage[capitalize]{cleveref}
\crefname{section}{Sec.}{Secs.}
\Crefname{section}{Section}{Sections}
\Crefname{table}{Table}{Tables}
\crefname{table}{Tab.}{Tabs.}

\def\cvprPaperID{7043} 
\def\confName{CVPR}
\def\confYear{2022}

\begin{document}

\title{Vision-Language Pre-Training for Boosting Scene Text Detectors}
\def\methodShort{VLPT-STD}


\author{
Sibo Song$^1$$^*$~~
Jianqiang Wan$^1$$^*$~~
Zhibo Yang$^1$~~
Jun Tang$^1$~~
Wenqing Cheng$^2$~~
Xiang Bai$^2$~~
Cong Yao$^1$
\smallskip
\\
$^1$DAMO Academy, Alibaba Group
\\
$^2$Huazhong University of Science and Technology
\smallskip
\\ 
\small{\texttt{\{sibosongzju,hustwjq,yangzhibo450,yaocong2010\}@gmail.com}} \\
\small{\texttt{xixing.tj@alibaba-inc.com}~~~~}
\small{\texttt{\{xbai,chengwq\}@hust.edu.cn}}
}

\maketitle

\def\thefootnote{*}\footnotetext{Both authors contributed equally to this work.}
\def\thefootnote{\arabic{footnote}}

\begin{abstract}
Recently, vision-language joint representation learning has proven to be highly effective in various scenarios. In this paper, we specifically adapt vision-language joint learning for scene text detection, a task that intrinsically involves cross-modal interaction between the two modalities: vision and language, since text is the written form of language. Concretely, we propose to learn contextualized, joint representations through vision-language pre-training, for the sake of enhancing the performance of scene text detectors. Towards this end, we devise a pre-training architecture with an image encoder, a text encoder and a cross-modal encoder, as well as three pretext tasks: image-text contrastive learning (ITC), masked language modeling (MLM) and word-in-image prediction (WIP). The pre-trained model is able to produce more informative representations with richer semantics, which could readily benefit existing scene text detectors (such as EAST and PSENet) in the down-stream text detection task. Extensive experiments on standard benchmarks demonstrate that the proposed paradigm can significantly improve the performance of various representative text detectors, outperforming previous pre-training approaches. The code and pre-trained models will be publicly released.
\end{abstract}

\vspace{-.5\baselineskip}
\section{Introduction}
\label{sec:intro}

\vspace{-.5\baselineskip}
\begin{figure}
  \centering
  \includegraphics[width=.86\linewidth]{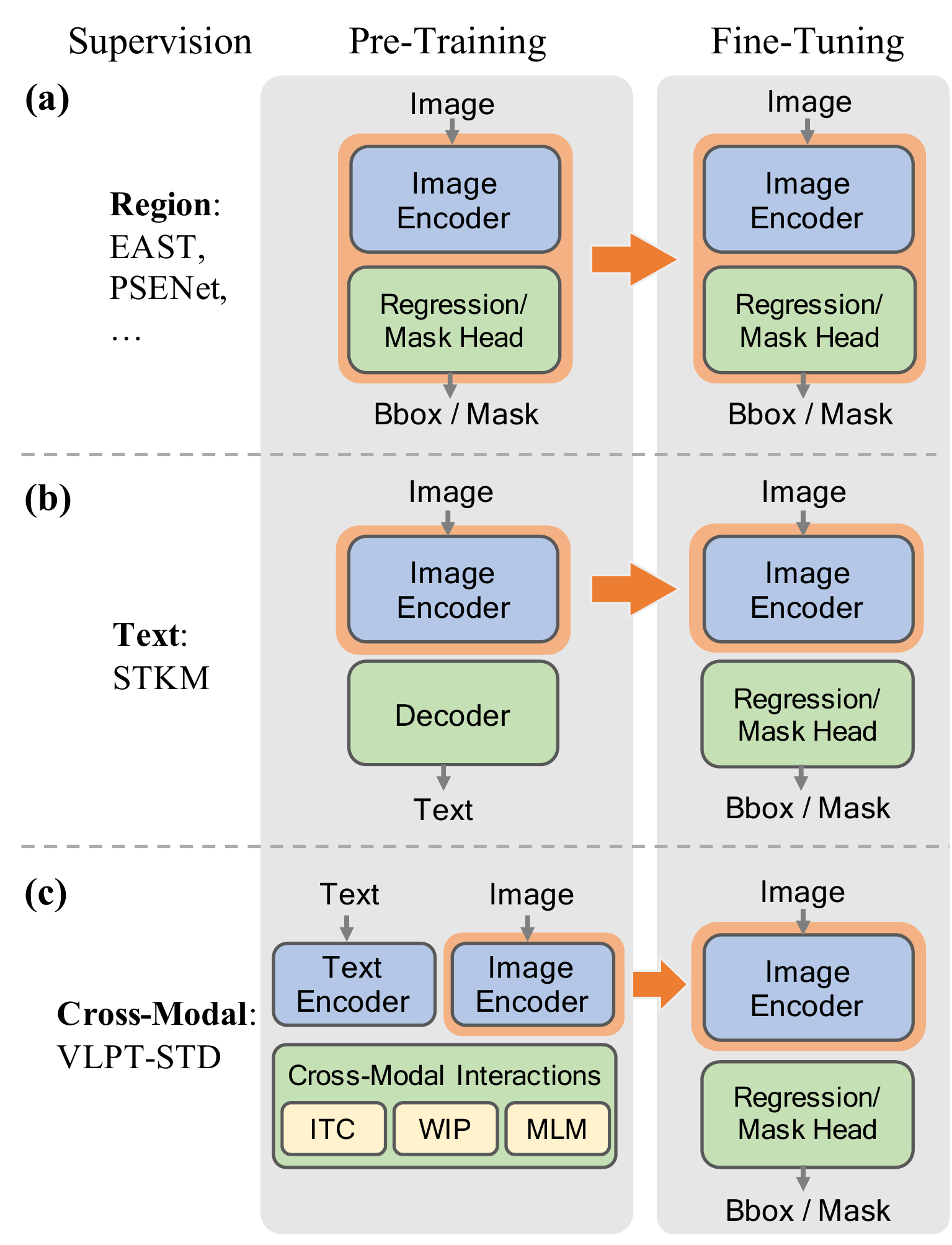}
  \vspace{-.5\baselineskip}
  \caption{Comparisons of pre-training paradigms for scene text detection. 
  \textbf{(a)} Conventional SynthText pre-training: Region supervisions (e.g., bounding box annotations, ground-truth masks) are used to train the image backbone and detector head. The fine-tuning pipeline is the same as the pre-training pipeline. 
  \textbf{(b)} Pre-training with text supervision: Text annotations are utilized as the supervisions through an encoder-decoder framework. 
  \textbf{(c)} Our pre-training model includes an image encoder, a text encoder and a cross-modal encoder, learned through three pretext tasks.}
  \label{fig:paradigm_compare}
  \vspace{-1.5\baselineskip}
\end{figure}

Scene text detection is a fundamental yet challenging task in computer vision, which requires the model to predict bounding boxes or polygons for each text instance in an image. For years, scene text detection methods based on deep learning have been extensively studied and widely adopted in both the academia and the industry due to its high research value and broad real-world applications. Recently, substantial advances have been observed, while grand challenges are still remained~\cite{long2021scene}.

Different pre-training strategies have been proposed to learn better representations in natural language processing~\cite{devlin2018bert} and computer vision~\cite{he2019rethinking}, usually relying on various pretext tasks.
To accelerate the training procedure and enhance the generalization capability, pre-training techniques have also largely been applied to scene text detection methods. Most of the early attempts employ ImageNet~\cite{deng2009imagenet} pre-training as done in general object detection. 
Nevertheless, an obvious domain gap exists between natural images from ImageNet and scene text images, which might result in limited performance gain after fine-tuning.

Therefore, researchers presented methods~\cite{he2021most,liao2020real} that fine-tune models pre-trained on synthetic text datasets, such as SynthText~\cite{gupta2016synthetic}, curved SynthText~\cite{long2019rethinking} and UnrealText~\cite{long2020unrealtext}. Most of recent text detection models that pre-trained on synthetic datasets outperform those pre-trained on ImageNet, however, they still suffer from a domain gap between synthetic and read-world data, which usually cause text-like textures to be falsely detected. 

To tackle this issue, Wan \etal ~\cite{wan2021self} propose STKM for pre-training via mining text knowledge without using region supervision. 
By adopting a text-recognition-like pipeline, STKM has proven to be effective on down-stream text detection task across different methods and datasets. 
However, STKM utilizes a character-level decoding process, which makes it hard to effectively exploit context information in the lexicon. In addition, the single-stream pipeline is essentially a unidirectional mapping (from vision modality to language modality), thus can not sufficiently make use of the interaction between vision and text to learn informative representations. 

We address these major challenges by introducing a novel \textbf{V}ision-\textbf{L}anguage \textbf{P}re-\textbf{T}raining paradigm for boosting \textbf{S}cene \textbf{T}ext \textbf{D}etectors, termed \textbf{\methodShort},
together with three novel pre-training objectives, which enable encoding richer information and learning discriminative representations. Importantly, by imposing a \textit{mutual alignment} between the two modalities, our method can better exploit text knowledge and achieve improved visual representations. 
As shown in \cref{fig:paradigm_compare}, we compare different pre-training paradigms for scene text detection. Note that, although our approach only requires image-level text annotations like STKM, it is designed from a very different perspective where fine-grained cross-modality interaction is adopted to align unimodal embeddings for learning better representations.

Inspired by vision-language pre-training approaches~\cite{chen2020uniter,li2021align,li2019visualbert,kim2021vilt}, we utilize self-attention and cross-attention modules to build a unified architecture together with three carefully designed pre-training objectives. The image and text unimodal representations are aligned first via contrastive learning, and then attend to fine-grained text regions via pre-training tasks of masked language model and word-in-image prediction. Consequently, the pre-trained backbone can be fine-tuned for various text detectors to significantly improve the detection performance. 

Specifically, the image (or text) embeddings are firstly extracted from an image (or text) encoder, and then fed to cross-attention blocks for fine-grained cross-modal interactions via various pre-training tasks. By designing various pre-training objectives, we encourage the encoder to attend text regions in image data from cross-modal cues. The whole model can be trained end-to-end and the visual backbone can be transferred to different text detectors. 
In addition, the proposed paradigm requires only image-level text annotations, whose labeling cost is much cheaper than conventional region annotations, especially for curve text labeling. 
Extensive experiments are conducted on various text detectors and datasets to demonstrate the effectiveness of the pre-trained backbone.

In summary, the main contributions of this paper are three-fold:
\begin{itemize}
  \vspace{-0.5\baselineskip}
  \item We propose a novel vision-language joint learning framework for pre-training the visual backbones of scene text detectors,
  which is a conceptually simple and flexible framework to enable \textit{mutual alignment} between visual and textual representations.
  \vspace{-0.5\baselineskip}
  \item We devise three pretext tasks to encourage fine-grained vision-language interactions. 
  In particular, a novel Word-in-Image Prediction (WIP) task is designed with hard example sampling strategy for learning discriminative representations. 
  \vspace{-0.5\baselineskip}
  \item Extensive experiments demonstrate the effectiveness of our approach.
  In particular, with three classical scene text detection methods: EAST, PSENet and DB, our approach has shown consistent and considerable improvements on five text detection datasets over conventional and STKM pre-training techniques.
\end{itemize}

\begin{figure*}
  \centering
  \includegraphics[width=0.86\linewidth]{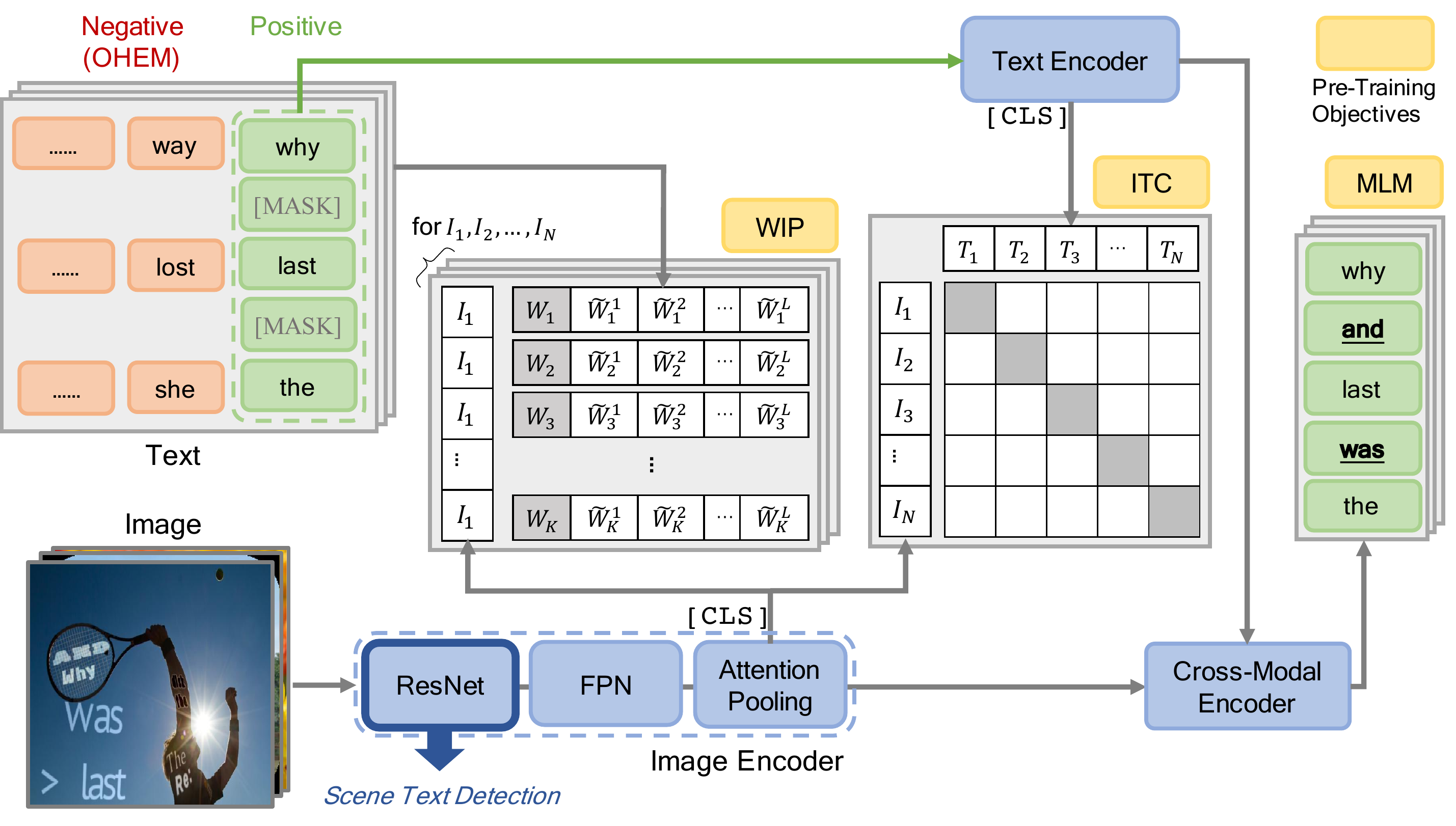}
  \vspace{-0.5\baselineskip}
  \caption{A schematic illustration of our proposed framework for cross-modal pre-training. Note that, only positive text are fed into text encoders for learning MLM task. }
  \label{fig:pipeline}
  \vspace{-1\baselineskip}
\end{figure*}

\section{Related Work}
\label{sec:related}

\noindent\textbf{Scene text detection.} Existing scene text detection methods can be roughly divided into two categories: bottom-up methods and top-down methods. 

Generally, bottom-up methods firstly detect fundamental components (e.g., pixels or segments), then aggregate the components to produce final detection results via various post-processing algorithms.
PixelLink~\cite{deng2018pixellink} firstly segments text instances with linking pixels and then generates bounding boxes from segmentation results. 
PSENet~\cite{li2018shape} firstly generates multiscale kernels for each text instance, and then progressively expands the minimal scale kernel to cover the whole text instance. 
SegLink and its variant SegLink++~\cite{shi2017detecting,tang2019seglink++} detect small text segments and then link them together to form the complete text instance.
CRAFT~\cite{baek2019character} detects the character region first and link the regions by learning the affinity between characters. 
DB~\cite{liao2020real} proposes a differentiable binarization module for a segmentation network.

Early attempts from top-down methods take scene text as general objects, apply object detection approaches~\cite{liu2016ssd,ren2015faster} to locate text regions by predicting the offsets from anchors or pixels. 
One-stage text detectors like TextBoxes series~\cite{liao2017textboxes,liao2018textboxes++}, EAST~\cite{zhou2017east}, MOST~\cite{he2021most} directly regress the geometry parameters of the text boxes on feature representations and apply Non-Maximum Suppression (NMS) algorithm to produce the final predictions. 
On the other hand, two-stage text detection methods like Mask TextSpotter series~\cite{lyu2018mask,liao2020mask}, generally follow a Mask R-CNN-style framework by utilizing a region proposal network (RPN) to produce text proposals first and then regress the offsets to the ground-truth bounding boxes.

\noindent\textbf{Pre-training of visual representations.} Despite the effective network design in previous works, pre-training techniques for scene text detection has not been sufficiently studied. 
Most of the existing works directly pre-train the network with region supervision on ImageNet or SynthText, without paying particular attention to the importance of pre-training techniques.

Modern pre-training approaches can drastically improve the performance on down-stream tasks for most deep learning applications, which has been observed for natural language processing~\cite{devlin2018bert}, computer vision~\cite{he2019rethinking} and cross domains~\cite{chen2020uniter,yang2021tap}.
Pre-training of visual representations has a long history with methods based on supervised training (from other data source)~\cite{he2019rethinking,ren2015faster}, self-supervised/unsupervised training~\cite{ranzato2007unsupervised,chen2020improved,chen2020simple,he2020momentum} and multimodal training~\cite{radford2021learning,yang2021tap,chen2020uniter}. 
Contrastive learning has become increasingly successful for self-supervised and multimodal pre-training. 
Methods like MoCo~\cite{he2020momentum} and SimCLR~\cite{chen2020simple} follow an instance discrimination pretext task, where the features of each image's multiple views are pulled away from those of other instances. 
The recent CLIP~\cite{radford2021learning}, ALIGN~\cite{jia2021scaling} and ALBEF~\cite{li2021align} perform pre-training on massive image-text data using contrastive loss by aligning the visual and textual representations. Performance on the down-stream tasks (e.g., image retrieval, image caption) is greatly improved with the pre-trained models. 

There has been an increasing interest in studying pre-training techniques for OCR-related approaches. TrOCR~\cite{li2021trocr} introduces an end-to-end text recognition method with a pre-trained Transformer-based model. 
Our work is closely related to STKM~\cite{wan2021self} which pre-trains ResNet backbone using high-level text knowledge with self-attention decoders. 
Importantly, unlike STKM, we propose a novel cross-modal pre-training paradigm inspired by the recent trends of vision-language pre-training techniques.
Instead of directly decoding text information from image representations, we demonstrate that learning better visual representations for boosting scene text detectors can be achieved by mutual alignment and cross-modal interactions through carefully designed pre-training objectives. 


\section{Methodology}
\label{sec:method}

In this section, we first present the overall paradigm of the proposed \methodShort, then describe the visual-linguistic pre-training tasks that are specifically designed to learn cross-modal representations for the down-stream task of text detection.

\subsection{Model Architecture}
\label{subsec:arch}

As illustrated in \cref{fig:pipeline}, \methodShort~contains an image encoder, a text encoder, and a cross-modal encoder. The image and text embeddings are first extracted from unimodal encoders, and then fed into a cross-modal encoder of several cross-attention layers, enabling fine-grained interactions across the image regions and the sub-word tokens. 

\noindent\textbf{Image Encoder.} The image encoder includes a ResNet~\cite{he2016deep} backbone, a feature pyramid network (FPN) and an attention pooling layer. 
For the first, to make fair comparisons with existing pre-trained models on the down-stream text detection task, a ResNet-50 is used as the base architecture for the feature extractor and initialized with weights pre-trained on ImageNet~\cite{deng2009imagenet}. 

Inspired by the success of FPN~\cite{2017Feature} on object detection, semantic segmentation, etc., we exploit feature fusion by combining features of $C_2$, $C_3$, $C_4$, and $C_5$ layers with lateral connections. 
Four convolutional feature maps are firstly transformed with number of channels reducing to 256 through an $1 \times 1$ convolution layer. 
Then $P_5$ is simply the transformed feature map of $C_5$, and ${P_2, P_3, P_4}$ are constructed by adding up-sampled transformed feature maps for the previous layer. 
Next, an $1 \times 1$ convolution layer with stride 2 is applied on the concatenated feature map to reduce the number of channels from 1024 to 384. 
Therefore, the feature map is $\frac{1}{16}$ of the original image size. 
Formally, the final feature map $\mathcal{F}_{c}$ is defined as,
\begin{equation}
  \resizebox{1.\hsize}{!}{%
  $ \mathcal{F}_{c} = \text{Conv}_{1 \times 1, s2} ([ \text{DS}_{\times 2}(P_2) ;  P_3 ; \text{US}_{\times 2}(P_4) ; \text{US}_{\times 4}(P_5) ]) $
  \label{eq:fpn}
  }
\end{equation}

where 
$\text{DS}_{\times x}(\cdot)$ denotes x times down-sampling and $\text{US}_{\times x}(\cdot)$ represent x times up-sampling with bilinear interpolation, respectively. 

In addition, we adopt an attention pooling mechanism as in \cite{radford2021learning} to extract visual embeddings conditioned on the global average-pooled representation of the image. 
The attention pooling block is implemented as one layer of Transformer's multi-head attention module. 

Overall, an input image $x^{I}$ is encoded into a sequence of embeddings $\mathbf{V} = \{ V_{\texttt{[CLS]}}, V_{1}, ..., V_{S} \} \in \mathbb{R}_{d}$, where $V_{\texttt{[CLS]}}$ denotes the embedding of the \texttt{[CLS]} token, $S$ denotes the number of visual tokens, and $d$ denotes the dimension of visual embeddings. 

\noindent\textbf{Text Encoder.} The text encoder consists of three multi-head self-attention modules which transforms an input text 
into a sequence of embeddings.

Given a text sample as input, we first split it into a sequence of words, and then apply WordPiece~\cite{sennrich2015neural} to tokenize each word into sub-word tokens. 
After that, an embedding matrix is adopted to embed the sub-word tokens into embedding vectors. 
Here we use $\mathbf{W} = \{W_{\texttt{[CLS]}}, W_{1}, W_{2}, \cdots, W_{K}\} \in \mathbb{R}_{d}$ to represent the embedding sequence, where $K$ indicates the sequence length, and $d$ is the dimension of word embeddings. 
Next, trainable positional embeddings are added following other BERT-based language models. 
Unlike most existing vision-language pre-trained models, the text encoder in our paradigm is trained from scratch, which is not initialized by any pre-trained BERT model. 

\noindent\textbf{Cross-Modal Encoder.} We construct four identical Transformer's decoder layers as the cross-modal encoder to enable interactions between visual and textual embeddings. Each decoder layer consists of a multi-head self-attention module, a multi-head cross-attention module (MHCA) and a feed-forward network (FFN).
Following~\cite{dosovitskiy2020image}, LayerNorm (LN) is applied \textit{before} every attention module, and residual connections \textit{after} every attention module. The FFN contains two MLP layers with a GELU~\cite{hendrycks2016gaussian} non-linearity. 
The last decoder layer has an additional prediction head for MLM task, which is described in \cref{subsec:task}.


\subsection{Pre-training Tasks}
\label{subsec:task}

We adopt three pretext tasks during pre-training: image-text contrastive learning (ITC) and word-in-image prediction (WIP) on the unimodal encoders, masked language modeling (MLM) on the cross-modal encoder. 

\noindent\textbf{Image-Text Contrastive learning (ITC).} Our goal is to learn visual unimodal representations that can be transferred for down-stream scene text detection task. 
Thus, the key challenge is to learn discriminative representations that can attend to text regions from the image encoder.

To achieve this, we employ the idea of contrastive learning to mutually align global textual and visual representations into semantic space.
ITC aims to find the best image embedding from a batch of image embeddings given a text embedding. 
Similarly, for a given image embedding, the objective is to find the best text embedding from a batch of text embeddings. 
In one word, the contrastive objective encourages the model to locate text in the image by jointly training the image and text encoders to maximize the cosine similarity of the paired image-text embeddings. 

Formally, we denote
each image-text pair from SynthText dataset as $(x^{I}_{i}, x^{T}_{i})$ which contains an image $x^{I}_{i}$ with the rendered text $x^{T}_{i}$ in the image. 
As previously illustrated, each image is encoded as image embedding $V_{\texttt{[CLS]}}$. 
Similarly, for each text, the text embedding $W_{\texttt{[CLS]}}$ is encoded by the text encoder. 
For notation simplicity, we use $I$ and $T$ to denote $V_{\texttt{[CLS]}}$ and $W_{\texttt{[CLS]}}$ respectively. 

The loss function for each data batch is constructed as follows: for each image query $x^{I}_{j}$, we obtain an InfoNCE~\cite{oord2018representation} loss between its image embedding ${I}_{j}$ and all text embeddings ${T}_{k}$ in the batch,
\vspace{-0.5\baselineskip}
\begin{equation}
  \mathcal{L}_{\text{I2T}} = -\sum_{j} \log \frac{ \exp \left( {I}_{j} \cdot {T}_{j} / \tau \right) }{ \sum_{k=1}^{N} \exp \left( {I}_{j} \cdot {T}_{k} / \tau \right) } 
  \label{eq:itc1}
\end{equation}

where $N$ denotes the batch size, and $\tau$ denotes the hyperparameter of temperature. The similarity is measured by dot product here. Similarly, for each text query $x^{T}_{j}$, the InfoNCE loss is formulated as,
\vspace{-0.5\baselineskip}
\begin{equation}
  \mathcal{L}_{\text{T2I}} = -\sum_{j} \log \frac{ \exp \left( {T}_{j} \cdot {I}_{j} / \tau \right) }{ \sum_{k=1}^{N} \exp \left( {T}_{j} \cdot {I}_{k} / \tau \right) } 
  \label{eq:itc2}
\end{equation}

The total loss function for ITC is defined as,
\vspace{-0.2\baselineskip}
\begin{equation}
  \mathcal{L}_{\text{ITC}} = \lambda_{1} \mathcal{L}_{\text{I2T}} + \lambda_{2} \mathcal{L}_{\text{T2I}}
  \label{eq:itc}
\end{equation}

Empirically, we find that $\lambda_{1}=\lambda_{2}=0.5$ work well in practice.

\noindent\textbf{Word-in-Image Prediction (WIP).} In addition to ITC, we propose a novel Word-in-Image Prediction (WIP) to enable fine-grained cross-modal interactions. 
In this task, we utilize contrastive learning between image embeddings and word-level embeddings to distinguish the positive words that rendered in the image from a number of negative words which are not in the image, therefore predicting its existence in the image. 

Inspired by the Online Hard Example Mining strategy (OHEM)~\cite{shrivastava2016training}, we further sample the hard negative samples based on text embeddings' similarities during training to capture more fine-grained visual cues from word shape. 
As illustrated in \cref{fig:pipeline}, for a positive word \textit{"lost"}, the sampled negative examples can be \textit{"last"}, \textit{"lose"}, \textit{"post"}, etc.
By virtue of OHEM and word-level contrastive learning, the visual representations will be iteratively improved through pre-training for modelling the subtle nuances between words of similar shape.
As shown in \cref{tab:text_embedding}, the learned text embeddings from our model indeed reveal visual similarity of word appearance via cross-modal alignment. 

Formally, we denote the positive sub-word tokens as $W = \{W_{\texttt{[CLS]}}, W_{1}, W_{2}, \cdots, W_{K}\} \in \mathbb{R}_{d}$, and its corresponding image embedding as $I$ for simplicity. For each token $W_{i}$, we sample its top $L$ nearest neighbors $\{\widetilde{W}_{i}^{1}, \widetilde{W}_{i}^{2}, \cdots, \widetilde{W}_{i}^{L}\}$ as negative examples in an online fashion, by measuring their word embeddings' similarities. 

Then the WIP objective function for each image-text pair $\{I, \{W_{1}, W_{2}, \cdots, W_{K}\} \}$ is constructed as,
\vspace{-0.2\baselineskip}
\begin{equation}
  \resizebox{1.\hsize}{!}{%
  $\mathcal{L}_{\text{WIP}} = -\sum_{k=1}^{K} \log \frac{ \exp \left( I \cdot W_{k} / \tau \right) }{ \exp \left( I \cdot W_{k} / \tau \right) + \sum_{l=1}^{L} \exp \left( I \cdot \widetilde{W}_{k}^{l} / \tau \right) } $%
  \label{eq:wip}
  }
\end{equation}

\noindent\textbf{Masked Language Modeling (MLM).} This objective is to predict the ground-truth labels of masked text tokens $W_{\text{masked}}$ from its contextualized vector $z_{\text{masked}} | W$. Following BERT~\cite{devlin2018bert}, we randomly mask $W$ with the probability of $0.15$ and the replacements are $10\%$ random tokens, $10\%$ unchanged, and $80\%$ {\tt [MASK]}. 

Specifically, the MLM task is to recover the masked word tokens based on the observation of their surrounding words $W_{\text{unmasked}}$ and all visual embeddings $\mathbf{V}$, by minimizing the negative log-likelihood,
\vspace{-0.2\baselineskip}
\begin{equation}
  \mathcal{L}_{\text{MLM}} = -\mathbb{E}_{({W}, {V})} \log P_{\theta}(W_{\text{masked}}|W_{\text{unmasked}},{\mathbf{V}})
  \label{eq:mlm}
\end{equation}

Note that the rendered text on SynthText images normally contains less semantically meaningful content unlike existing tasks of image caption~\cite{kim2021vilt,chen2020uniter} or document understanding~\cite{xu2020layoutlm,xu2020layoutlmv2}.
Therefore, the MLM task in our paradigm highly relies on the image content instead of the language context to recover the masked tokens. 
Thus, the model can learn stronger representations on the visual side for pre-training text detection task.

Overall, the full pre-training objective is,
\vspace{-0.2\baselineskip}
\begin{equation}
  \mathcal{L} = \mathcal{L}_{\text{ITC}} + \mathcal{L}_{\text{WIP}} + \mathcal{L}_{\text{MLM}}
  \label{eq:total}
\end{equation}

\vspace{-1.\baselineskip}
\section{Experiments}
\label{sec:experiments}

In this session, we first briefly introduce the SynthText dataset for pre-training and other publicly available datasets for scene text detection task. 
Then, we present the technical details for both pre-training and fine-tuning.
Next, we transfer the ResNet backbone from our \methodShort~to three classical text detection methods: EAST, PSENet and DB, and compare them with state-of-the-art pre-training methods on various challenging benchmarks. 
Finally, we conduct ablation studies and present qualitative results to demonstrate the effectiveness and generalization ability of our method.

\subsection{Experimental Settings}


\noindent\textbf{Datasets.} The pre-training experiments are conducted on SynthText \cite{gupta2016synthetic}, which is a large-scale dataset that contains about $800K$ synthetic images. 
We hold out a subset of $20K$ images for validation in the following experiments.
SynthText has a variety of labeling forms, we only use text labels for pre-training without any bounding box annotations. 
For down-stream experiments, we closely follow STKM~\cite{wan2021self} and use the following datasets for evaluation. 


Total-Text~\cite{ch2017total} mainly focuses on curved text, which contains 1,255 training images and 300 test images. The dataset is labeled with polygon-shaped bounding boxes.

CTW1500~\cite{yuliang2017detecting} also primarily consists of curved text. It has 1,000 training images and 500 test images. Text instances are labeled with polygons annotated by 14 vertices.


ICDAR2015~\cite{karatzas2015icdar} is composed of 1,000 training images and 500 test images. Text annotations are given in word level with rectangular bounding boxes.


ICDAR2017~\cite{nayef2017icdar2017} is a multi-lingual dataset including 9 different languages, which contains 7,200 training images, 1,800 validation images, and 9,000 test images. Both training set and validation set are used during fine-tuning.


MSRA-TD500~\cite{yao2012detecting} includes 300 training images and 200 test images with line-level annotations. Following previous works, 400 training images of HUST-TR400 \cite{yao2014unified} are used as additional training data.

TextOCR ~\cite{singh2021textocr} is a large and diverse OCR dataset that consists of 28,134 images and 903K annotated words in total, with high density of ${\sim}32$ words per image.

\begin{table}[]
  \centering
  \caption{Experimental results on PSENet. \textbf{ST} denotes SynthText pre-training. The numbers on $\Delta$ denote the F-measure improvements over SynthText and STKM in \textbf{\textcolor{Green}{green}} and \textbf{\textcolor{Blue}{blue}} respectively. }
  \vspace{-0.5\baselineskip}
  \scalebox{0.66}{
  \begin{threeparttable}
    \begin{tabular}{l|ccc|ccc|ccc}
    \toprule
      Methods  & \multicolumn{3}{c|}{ICDAR2015}  & \multicolumn{3}{c|}{Total-Text}  & \multicolumn{3}{c}{CTW1500}  \\
      & P & R & F & P & R & F & P & R & F  \\ \midrule
    SegLink~\cite{shi2017detecting}      &  73.1  &  76.8  &  75.0  &  30.3  &  23.8  &  26.7  &  42.3  &  40.0  &  40.8  \\ 
    TextSnake~\cite{long2018textsnake}   &  84.9  &  80.4  &  82.6  &  82.7  &  74.5  &  78.4  &  67.9  &  85.3  &  75.6  \\ 
    TextDragon~\cite{feng2019textdragon} &  84.8  &  81.8  &  83.1  &  79.5  &  81.0  &  80.2  &  84.5  &  74.2  &  79.0  \\
    SAE~\cite{tian2019learning}          &  84.5  &  85.1  &  84.8  &   -    &   -    &   -    &  82.7  &  77.8  &  80.1  \\
    \midrule
    PSENet + ST $^1$   & 84.3     & 78.4  & 81.3     & 89.2     & 79.2  & 83.9     & 83.6     & 79.7  & 81.6     \\
    PSENet + STKM $^1$ & 85.7     & 81.8  & 83.7     & 89.2     & 79.9  & 84.3     & 85.3     & 80.6  & 82.9     \\ 
    PSENet + Ours & 86.0     & 82.8  & \textbf{84.3} & 90.8  & 82.0  & \textbf{86.1} & 86.3     & 80.7     & \textbf{83.3}   \\ 
    $\Delta$   &  \multicolumn{3}{r|}{\textbf{\textcolor{Green}{3.0$\uparrow$}, \textcolor{Blue}{0.6$\uparrow$}}}  & \multicolumn{3}{r|}{\textbf{\textcolor{Green}{2.2$\uparrow$}, \textcolor{Blue}{1.8$\uparrow$}}}  & \multicolumn{3}{r}{\textbf{\textcolor{Green}{1.7$\uparrow$}, \textcolor{Blue}{0.4$\uparrow$}}}  \\ 
    \bottomrule
    \end{tabular}
    \begin{tablenotes}
      \item[1] We report results using our reimplementation.
    \end{tablenotes}
  \end{threeparttable}}
  \label{tab:pse_results}
  \vspace{-0.5\baselineskip}
  \end{table}
  
  \begin{table}[]
  \centering
  \caption{Experimental results on EAST. \textbf{ST} denotes SynthText pre-training. The numbers on $\Delta$ denote the F-measure improvements over SynthText and STKM in \textbf{\textcolor{Green}{green}} and \textbf{\textcolor{Blue}{blue}} respectively. }
  \vspace{-0.5\baselineskip}
  \scalebox{0.67}{
  \begin{threeparttable}
  \begin{tabular}{l|ccc|ccc|ccc}
  \toprule
  Methods  & \multicolumn{3}{c|}{ICDAR2015}  & \multicolumn{3}{c|}{ICDAR2017}  & \multicolumn{3}{c}{MSRA-TD500}  \\
  & P & R & F & P & R & F & P & R & F  \\ \midrule
  SegLink~\cite{shi2017detecting} &  73.1  &  76.8  &  75.0  &  -  &  -  &  -  &  86  &  70  &  77  \\ 
  TextField~\cite{xu2019textfield}  &  84.3  &  80.1  &  82.4  &  -  &  -  &  -  &  87.4  &  75.9  &  81.3  \\ 
  CRAFT~\cite{baek2019character}   &  89.8  &  84.3  &  86.9  &  80.6 &  68.2  &  73.9  &  88.2  &  78.2  &  82.9  \\
  GNNets~\cite{xu2019geometry}  &  90.4  &  86.7  &  88.5  &  79.6 &  70.1  &  74.5  &  -     &  -     &  -  \\
  \midrule
  EAST + ST $^1$ & 89.6 & 81.5 & 85.3 & 75.1 & 61.9 & 67.9 & 86.9 & 77.6 & 82.0     \\
  EAST + STKM $^1$     & 90.2 & 84.6 & 87.3 & 76.9 & 64.3 & 70.0 & 85.2 & 75.3 & 80.0     \\ 
  EAST + Ours      & 91.5 & 85.4 & \textbf{88.3} & 77.7 & 64.6  & \textbf{70.5}  & 88.5     & 76.7  & \textbf{82.2}    \\ 
  $\Delta$   &  \multicolumn{3}{r|}{\textbf{\textcolor{Green}{3.0$\uparrow$}, \textcolor{Blue}{1.0$\uparrow$}}}  & \multicolumn{3}{r|}{\textbf{\textcolor{Green}{2.6$\uparrow$}, \textcolor{Blue}{0.5$\uparrow$}}}  & \multicolumn{3}{r}{\textbf{\textcolor{Green}{0.2$\uparrow$}, \textcolor{Blue}{2.2$\uparrow$}}}  \\ 
  \bottomrule
  \end{tabular}
  \begin{tablenotes}
    \item[1] We report results using our reimplementation.
  \end{tablenotes}
  \end{threeparttable}}
  \label{tab:east_results}
  \vspace{-0.5\baselineskip}
  \end{table}
  
  \begin{table}[]
    \centering
    \caption{Experimental results on DB. \textbf{ST} denotes SynthText pre-training. The numbers on $\Delta$ denote the F-measure improvements over SynthText and STKM in \textbf{\textcolor{Green}{green}} and \textbf{\textcolor{Blue}{blue}} respectively. }
    \vspace{-0.5\baselineskip}
    \scalebox{0.69}{
    \begin{threeparttable}
    \begin{tabular}{l|ccc|ccc|ccc}
    \toprule
      Methods  & \multicolumn{3}{c|}{ICDAR2015}  & \multicolumn{3}{c|}{Total-Text}  & \multicolumn{3}{c}{MSRA-TD500}  \\ 
    &  P & R & F & P & R & F & P & R & F  \\ \midrule
    DB + ST $^1$    & 88.2 & 82.7 & 85.4 & 87.1 & 82.5 & 84.7 & 91.5 & 79.2 & 84.9     \\
    DB + STKM $^1$  & 91.4 & 81.4 & 86.1 & 87.7 & 83.4 & 85.5 & 90.2 & 82.0 & 85.9     \\ 
    DB + Ours      & 92.0 & 81.6 & \textbf{86.5} & 88.7 & 84.0  & \textbf{86.3}  & 92.3     & 84.9  & \textbf{88.5}    \\ 
    $\Delta$   &  \multicolumn{3}{r|}{\textbf{\textcolor{Green}{1.1$\uparrow$}, \textcolor{Blue}{0.4$\uparrow$}}}  & \multicolumn{3}{r|}{\textbf{\textcolor{Green}{1.6$\uparrow$}, \textcolor{Blue}{0.8$\uparrow$}}}  & \multicolumn{3}{r}{\textbf{\textcolor{Green}{3.6$\uparrow$}, \textcolor{Blue}{2.6$\uparrow$}}}  \\ 
    \bottomrule
    \end{tabular}
    \begin{tablenotes}
      \item[1] We report results using our reimplementation.
    \end{tablenotes}
    \end{threeparttable}}
    \label{tab:db_results}
    \vspace{-1.0\baselineskip}
  \end{table}

\noindent\textbf{Pre-training setup.} During pre-training, the SynthText images are first resized to $512 \times 512$ and then rotated randomly within the range of \ang{-20} to \ang{+20}, other data augmentation strategies are not included due to the lack of region supervisions.
We adopt AdamW optimizer~\cite{loshchilov2017decoupled} with 
weight decay of $0.01$ and warm-up learning rate schedule for first $2.5K$ steps, then the learning rate was linearly decayed from $1 \times 10^{-4}$ to $0$. The model is trained for $120K$ iterations on 8 Tesla V100 GPUs with a batch size of 800. 
To save memory and accelerate the pre-training process, we apply mixed-precision~\cite{micikevicius2017mixed} and gradient checkpointing~\cite{griewank2000algorithm} techniques.
The learnable temperature parameter $\tau$ in \cref{eq:itc1}, \cref{eq:itc2} and \cref{eq:wip} are initialized to $0.07$ and clipped to prevent scaling the logits by larger than $100$ to improve training stability.
We set the dimension $d$ of visual and text embedding to $384$. 
The number of visual tokens $S$ and textual tokens $K$ are set to $1025$ and $30$ respectively. 
We select $L=63$ most similar word tokens as the hard negative examples. 

\noindent\textbf{Evaluation protocol.} We evaluate the pre-trained model by fine-tuning EAST, PSENet and DB methods on five commonly used scene text detection datasets. 

For EAST method, ICDAR2015, ICDAR2017 and MSRA-TD500 are used as benchmark datasets as it is not suitable for detecting curved texts.
We follow ~\cite{east2022github} and employ a variety of data augmentations, including horizontal flips, rotation, resizing, random crops and color jittering over input images 
During training, the images of ICDAR2015, ICDAR2017, and MSRA-TD500 are resized to $512 \times 512$, $640 \times 640$, $640 \times 640$, and the batch size is $32$, $22$, and $22$, respectively. 
We fine-tune the pre-trained backbone for 600 epochs on ICDAR2015 and MSRA-TD500 datasets with Adam optimizer, base learning rate is $1 \times 10^{-4}$ and decayed by $0.1$ every $200$ epochs. For ICDAR2017, EAST is trained for $300$ epochs and the initial learning rate is $1 \times 10^{-4}$ then decayed by $0.1$ every $50$ epochs.

For PSENet, we use the official code ~\cite{pse2022github} and conduct experiments on ICDAR2015, Total-Text and CTW1500.
The experimental settings follow the original paper~\cite{li2018shape}.

For DB, we adopt the official implementation ~\cite{db2022github} to reproduce the baseline and evaluate our pre-trained method.
The model is trained for $1200$ epochs following official code. 
We adopt Adam optimizer with a weight decay of $0.0001$ and an initial learning rate of $1 \times 10^{-4}$ which is decreased by $0.1$ after epoch $800$.
For other parameters, we directly follow the default settings in ~\cite{db2022github}.

\begin{table}[]
  \centering
  \caption{ Architecture designs. $\dagger$ denotes the original FPN implementation from \cite{2017Feature}. MHCA denotes the multi-head cross-attention. Only F-measure is presented.}
  \vspace{-0.5\baselineskip}
  \scalebox{.84}{
  \begin{tabular}{cc|cc|cc}
  \toprule
  \multicolumn{2}{c|}{Image Encoder} & \multicolumn{2}{c|}{Cross-Modal Encoder} & PSENet & EAST \\
  \midrule
  $\text{FPN}^{\dagger}$ & Our FPN  & w/o MHCA & w/ MHCA & CTW    & IC15    \\
  \midrule
   $\surd$  &          &          & $\surd$  &  82.7  &  88.0  \\
            & $\surd$  & $\surd$  &          &  83.1  &  87.4  \\
            & $\surd$  &          & $\surd$  &  83.3  &  88.3  \\
  \bottomrule
  \end{tabular}}
  \label{tab:ablation_arch}
  \vspace{-1.\baselineskip}
\end{table}

\begin{table}[]
\centering
\caption{ Ablation study for \methodShort~ pre-training tasks. Only F-measure is presented. }
\vspace{-0.5\baselineskip}
\scalebox{.79}{
  \begin{tabular}{ccc|ccc|ccc}
  \toprule
           &         &         & \multicolumn{3}{c|}{PSENet} & \multicolumn{3}{c}{EAST} \\
     ITC   &   MLM   &   WIP   &  IC15  &  TT    &  CTW   &  IC15  &  IC17  &  TD500  \\
  \midrule
           &         &         &  81.3  &  83.9  &  81.6  &  85.3  &  67.9  &  82.0   \\
  \midrule
   $\surd$ &         &         &  82.2  &  84.3  &  82.2  &  86.0  &  69.4  &  79.3   \\
           & $\surd$ &         &  84.5  &  85.9  &  83.1  &  87.7  &  70.2  &  81.5   \\
           &         & $\surd$ &  83.1  &  85.3  &  82.2  &  86.9  &  70.2  &  82.1   \\
  \midrule
   $\surd$ & $\surd$ &         &  84.3  &  85.6  &  83.2  &  87.5  &  70.3  &  81.7   \\
   $\surd$ &         & $\surd$ &  83.3  &  85.3  &  82.5  &  87.3  &  70.2  &  81.5   \\
           & $\surd$ & $\surd$ &  84.7  &  85.8  &  82.9  &  87.6  &  70.4  &  81.9   \\
  \midrule
   $\surd$ & $\surd$ & $\surd$ &  84.3  &  86.1  &  83.3  &  88.3  &  70.5  &  82.2   \\
  \bottomrule
  \end{tabular}}
\label{tab:ablation_task}
\vspace{-1.\baselineskip}
\end{table}

\begin{table}[]
  \centering
  \caption{ Ablation study on pre-training datasets with PSENet. \textbf{ST} denotes SynthText and \textbf{TO} denotes TextOCR. Only F-measure is presented. }
  \vspace{-0.5\baselineskip}
  \scalebox{.90}{
  \begin{tabular}{c|c|c|c|c}
  \toprule
  Supervision & Pre-training Datasets & IC15 & TT & CTW \\ \midrule
  Region      & ST+TO (1 epoch) & 82.24 & 84.52 & 81.75  \\ 
  Text (Ours) & ST (1 epoch)    & 84.33 & 86.14 & 83.30  \\
  Text (Ours) & ST+TO (1 epoch) & 85.07 & 86.18 & 83.50  \\
  \bottomrule
  \end{tabular}}
  \label{tab:texocr_results}
  \vspace{-1.\baselineskip}
\end{table}

\begin{table*}[]
\centering
\caption{ Comparisons of top-5 nearest neighbors of sampled query words from our text encoder and ViLT's text encoder. }
\vspace{-0.5\baselineskip}
\scalebox{.84}{
\begin{tabular}{l|lllll|lllll}
\toprule
Query  & \multicolumn{5}{c|}{Top-5 nearest neighbors from \methodShort~} & \multicolumn{5}{c}{Top-5 nearest neighbors from ViLT\cite{kim2021vilt}}                            \\ 
\midrule
\texttt{eco}  & 850  & 800  & 630   & rca   & 600  & ecological  & dairy & organic   & international       & retro      \\
\texttt{vote}  & note   & voice  & work   & role   & write & voting & qualify & election & comment   & compete  \\ 
\texttt{sale} & safe  & scale  & said  & able   & sake & rent & display    & adoption     & auction      & hire  \\
\texttt{north} & worth  & keith  & math   & norton & both  & south  & east    & west     & northeast & southeast \\
\texttt{river} & liver & layer & viper & driver & meter & lake & creek & canal & rivers & waterway \\
\texttt{right} & light & night & rights & might & higher & left & starboard & back & sideways & bottom  \\
\texttt{special} & specific & typical & serial & social & optical & private & important & wonderful & new & significant \\
\texttt{affected} & attached & selected & attacked & scattered & affiliated & damaged & threatened & caused & involved & killed  \\
\bottomrule
\end{tabular}}
\label{tab:text_embedding}
\vspace{-1.\baselineskip}
\end{table*}



\subsection{Comparisons with State-of-the-Arts}

In \cref{tab:pse_results}, \cref{tab:east_results} and \cref{tab:db_results}, we compare the proposed \methodShort~with conventional SynthText pre-training and the state-of-the-art STKM method. 
Extensive experiments are conducted on five scene text detection datasets with PSENet, EAST and DB methods.
P, R, and F denote Precision, Recall and F-measure, respectively.
For reported STKM results, we use pre-trained model\footnote{\url{https://github.com/CVI-SZU/STKM}} released by the authors of \cite{wan2021self}.

Note that, there is no deformable convolutional layer in either STKM or our model. 
Thus, when fine-tuning DB with STKM's or our pre-trained backbone, the weights of deformable convolutional layers are randomly initialized.

Our method has shown consistent improvements over conventional SynthText pre-training and STKM on all benchmarking datasets, including two datasets for curved text, one dataset for multi-oriented scene text, and two multi-lingual datasets for long text lines.

\subsection{Ablation Study}

\noindent\textbf{Evaluation on architecture designs.} We compare different architecture designs for image encoder and cross-modal encoder in \cref{tab:ablation_arch}. 

For FPN, we compare the original FPN implementation~\cite{2017Feature} that used in \cite{li2018shape}. 
Our FPN variant reduces training computation and improves fine-tuning accuracy by 0.6\% and 0.3\% on CTW1500 and ICDAR2015, respectively.
In addition, we also study the cross-attention mechanism in cross-modal encoder design. 
Cross-modal encoder implemented with only self-attention modules leads to a fine-tuning performance drop, which indicates the superiority of the  cross-attention modules. 

\noindent\textbf{Evaluation on pre-training tasks.} To investigate the effectiveness of different pre-training objectives, we conduct ablative experiments with PSENet and EAST, and report the results in \cref{tab:ablation_task}. 

We find that pre-training either with MLM or WIP task boosts the accuracy significantly. 
We presume that the MLM and WIP design concepts have a conceptual advantage, \textit{utilizing word-level tokens for fine-grained cross-modal alignments}. 
Despite the superior performance of MLM and WIP, pre-trained model with single ITC task achieves higher performance than SynthText pre-training except for MSRA-TD500 dataset, suggesting that the global contrastive learning between image and text modalities is also effective in learning visual representations. 
While ITC + MLM already sets a strong baseline, integrating WIP brings extra performance boost.
Moreover, if all three pretext tasks are utilized, our approach consistently improves over the baseline by a nontrivial margin on all five datasets. 

\noindent\textbf{Evaluation on pre-training datasets.} Our method is compatible with both synthetic and real image datasets.
Therefore, we ablate pre-training datasets and experiment with TextOCR (${\sim}20k$) and SynthText (${\sim}800k$) in \cref{tab:texocr_results}. 
When combining SynthText and TextOCR as the pre-training dataset, it can outperform classical pre-training method using region supervision. 
Furthermore, it achieves a noticeable boost over SynthText-only pre-training on ICDAR15 which contains many small scene text instances and thus is more challenging. 

\subsection{Qualitative Results}
\label{subsubsec:qualitative}

\begin{figure*}
  \centering
  \includegraphics[width=.97\linewidth]{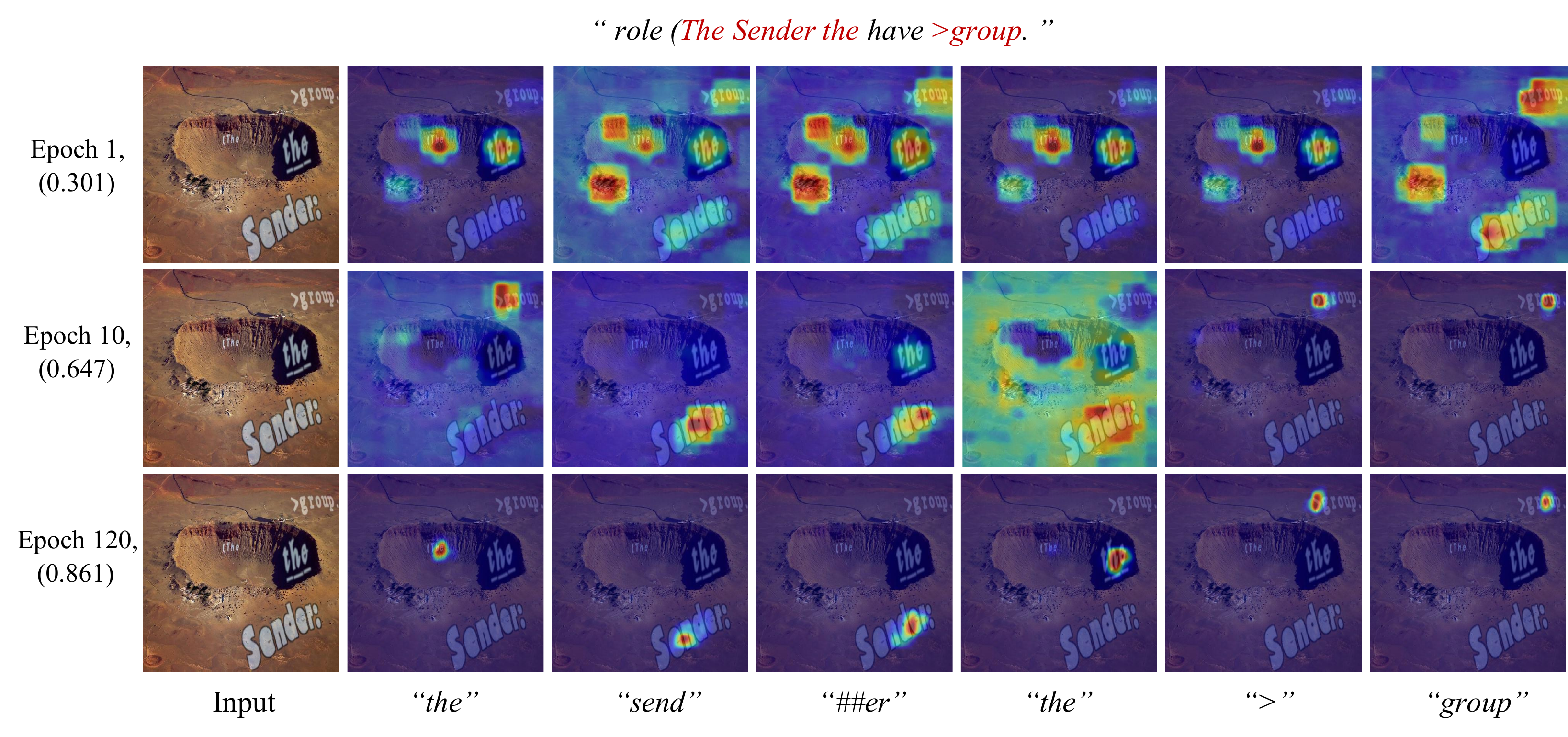}
  \vspace{-1.\baselineskip}
  \caption{Visualizations on the cross-attention maps corresponding to individual sub-word tokens. We present the visualization from the 1st head in the 2nd block of the cross-modal encoder. The selected sub-word tokens are highlighted in \textbf{\textcolor{Red}{red}}. The numbers in the brackets indicate the MLM accuracy on the validation set. (The figure is best viewed in color.)}
  \label{fig:attn}
  \vspace{-1.\baselineskip}
\end{figure*}

\noindent\textbf{Visualizations of attention maps.} We provide the per-word attention map visualizations from our cross-modal encoder in \cref{fig:attn}. 
As can be seen in the visualizations from the 1st training epoch, the model has acquired the ability to locate general text regions. 
While, it cannot capture fine-grained visual cues to distinguish different words at the beginning of the pre-training.
As training proceeds, the model gradually learns to attend to the corresponding text regions in the image for different words. 
As shown in the last row, the pre-trained model succeeds in learning accurate attention via the cross-modal encoder at the end of pre-training. 
More interestingly, for the two same \textit{"the"} input tokens, the attention mechanism enables the model to collaborate for attending different text regions. 

\begin{figure}
  \centering
  \includegraphics[width=0.91\linewidth]{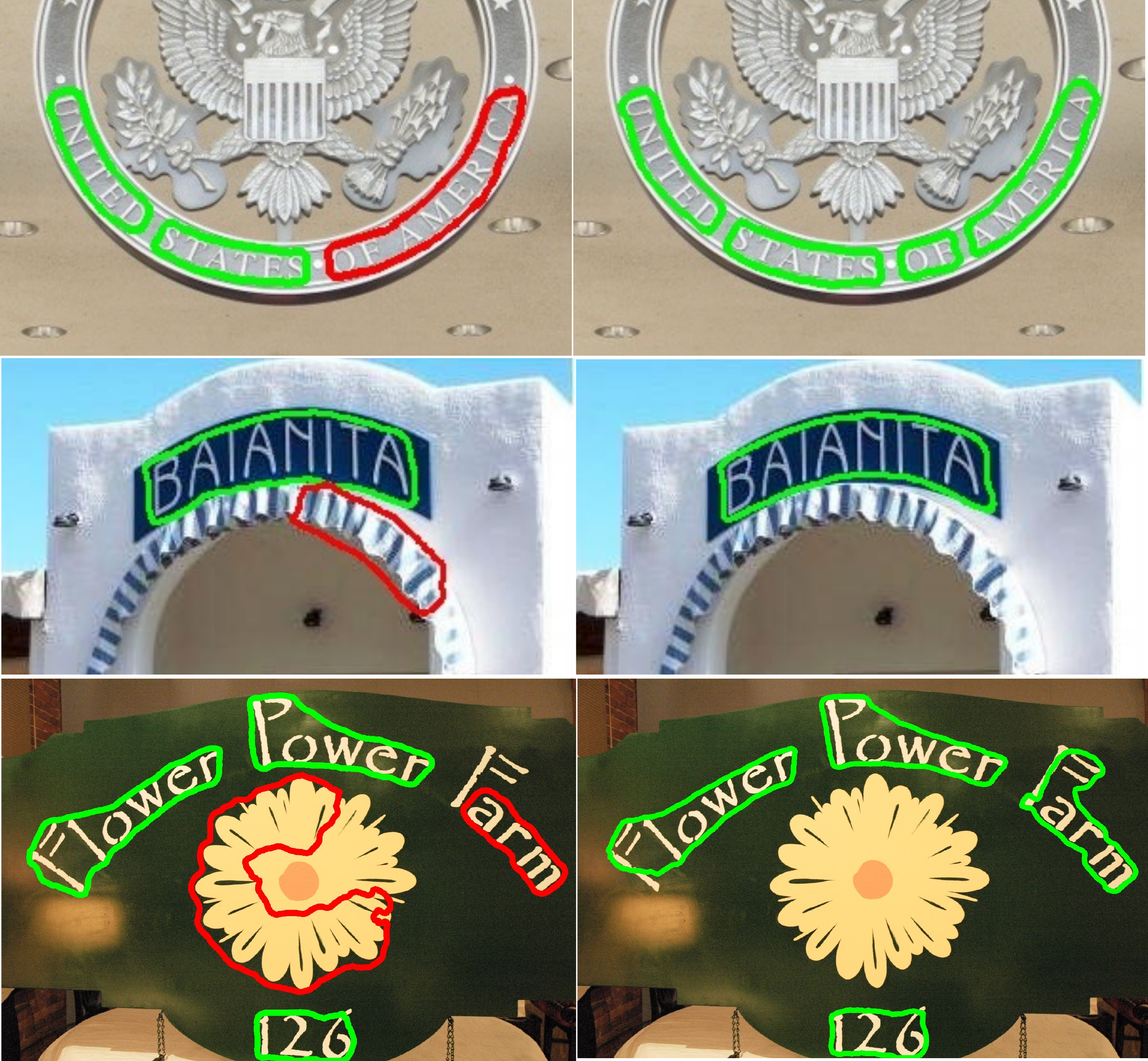}
  \vspace{-0.5\baselineskip}
  \caption{Visualizations of text detection results from STKM (\textbf{Left}) and \methodShort~(\textbf{Right}) on Total-Text test images after fine-tuning with PSENet. (The figure is best viewed in color.)}
  \label{fig:tt_compare}
  \vspace{-1.\baselineskip}
\end{figure}

\noindent\textbf{Visualizations of detection results.} \cref{fig:tt_compare} compares the detection results from STKM and our method \textit{after} fine-tuning with PSENet. 
As shown in the figure, our model can further suppress the false detections on text-like regions compared to STKM, as we apply MLM and WIP to enhance fine-grained representations by multimodal cues. 
Moreover, by using text encoder with WordPiece tokenization, our \methodShort~model can benefit from the integration of lexicon information in improving the segmentation quality in terms of both granularity and integrity, like \textit{"OF AMERICA"} in the first row and \textit{"Farm"} in the third row. 

\noindent\textbf{Nearest neighbors of word embeddings.} As depicted in \cref{tab:text_embedding}, we show top-5 nearest neighbors for sampled query words according to cosine similarity of embedding vectors from our pre-trained text encoder and text encoder of ViLT \cite{kim2021vilt}. 
As expected, the nearest neighbors generated by ViLT text encoder are quite semantically similar. 
While, the nearest neighbors from ours are more \textit{visually} similar to each other. 
This observation indicates that the mutual alignment between multimodal representations is achieved, which further validates the effectiveness of our cross-modal pre-training paradigm. 



\vspace{-0.5\baselineskip}
\section{Limitations}
\label{sec:limitations}

Our proposed method pre-trains text detection backbones with extra modules, e.g., text encoder and cross-modal encoder, comparing to classical text detection pre-training methods. 
Therefore, the whole pipeline consumes larger GPU memory and takes longer time for pre-training.
Moreover, if our approach is applied to other datasets where \textit{significantly} more words exist, e.g., document images of thousands of words, the number of text tokens will increase greatly. 
It will take higher computation and memory cost, which might not be environmentally friendly. 

\section{Conclusions and Future Works}

In this paper, we present \methodShort, a novel pre-training paradigm that contains unimodal encoders and a cross-modal encoder for vision-language joint learning. 
Three pre-training objectives are proposed to encourage fine-grained mutual alignments between image and text modalities, enabling visual encoder to incorporate information from lexicons for improving down-stream text detection task. 
Trained with five commonly used datasets on two representative text detection methods, our proposed approach outperforms state-of-the-art pre-training methods by a significant margin. 

Future works include integrating region supervision in the proposed paradigm, as well as developing more effective pre-training tasks.

{\small
\bibliographystyle{ieee_fullname}
\bibliography{egbib}

\begin{thebibliography}{10}\itemsep=-1pt

\bibitem{baek2019character}
Youngmin Baek, Bado Lee, Dongyoon Han, Sangdoo Yun, and Hwalsuk Lee.
\newblock Character region awareness for text detection.
\newblock In {\em Proceedings of the IEEE/CVF Conference on Computer Vision and
  Pattern Recognition}, pages 9365--9374, 2019.

\bibitem{chen2020simple}
Ting Chen, Simon Kornblith, Mohammad Norouzi, and Geoffrey Hinton.
\newblock A simple framework for contrastive learning of visual
  representations.
\newblock In {\em International conference on machine learning}, pages
  1597--1607. PMLR, 2020.

\bibitem{chen2020improved}
Xinlei Chen, Haoqi Fan, Ross Girshick, and Kaiming He.
\newblock Improved baselines with momentum contrastive learning.
\newblock {\em arXiv preprint arXiv:2003.04297}, 2020.

\bibitem{chen2020uniter}
Yen-Chun Chen, Linjie Li, Licheng Yu, Ahmed El~Kholy, Faisal Ahmed, Zhe Gan, Yu
  Cheng, and Jingjing Liu.
\newblock Uniter: Universal image-text representation learning.
\newblock In {\em European conference on computer vision}, pages 104--120.
  Springer, 2020.

\bibitem{ch2017total}
Chee~Kheng Ch'ng and Chee~Seng Chan.
\newblock Total-text: A comprehensive dataset for scene text detection and
  recognition.
\newblock In {\em 2017 14th IAPR International Conference on Document Analysis
  and Recognition (ICDAR)}, volume~1, pages 935--942. IEEE, 2017.

\bibitem{deng2018pixellink}
Dan Deng, Haifeng Liu, Xuelong Li, and Deng Cai.
\newblock Pixellink: Detecting scene text via instance segmentation.
\newblock In {\em Proceedings of the AAAI Conference on Artificial
  Intelligence}, volume~32, 2018.

\bibitem{deng2009imagenet}
Jia Deng, Wei Dong, Richard Socher, Li-Jia Li, Kai Li, and Li Fei-Fei.
\newblock Imagenet: A large-scale hierarchical image database.
\newblock In {\em 2009 IEEE conference on computer vision and pattern
  recognition}, pages 248--255. Ieee, 2009.

\bibitem{devlin2018bert}
Jacob Devlin, Ming-Wei Chang, Kenton Lee, and Kristina Toutanova.
\newblock Bert: Pre-training of deep bidirectional transformers for language
  understanding.
\newblock {\em arXiv preprint arXiv:1810.04805}, 2018.

\bibitem{dosovitskiy2020image}
Alexey Dosovitskiy, Lucas Beyer, Alexander Kolesnikov, Dirk Weissenborn,
  Xiaohua Zhai, Thomas Unterthiner, Mostafa Dehghani, Matthias Minderer, Georg
  Heigold, Sylvain Gelly, et~al.
\newblock An image is worth 16x16 words: Transformers for image recognition at
  scale.
\newblock In {\em International Conference on Learning Representations}, 2020.

\bibitem{feng2019textdragon}
Wei Feng, Wenhao He, Fei Yin, Xu-Yao Zhang, and Cheng-Lin Liu.
\newblock Textdragon: An end-to-end framework for arbitrary shaped text
  spotting.
\newblock In {\em Proceedings of the IEEE/CVF International Conference on
  Computer Vision}, pages 9076--9085, 2019.

\bibitem{griewank2000algorithm}
Andreas Griewank and Andrea Walther.
\newblock Algorithm 799: revolve: an implementation of checkpointing for the
  reverse or adjoint mode of computational differentiation.
\newblock {\em ACM Transactions on Mathematical Software (TOMS)}, 26(1):19--45,
  2000.

\bibitem{gupta2016synthetic}
Ankush Gupta, Andrea Vedaldi, and Andrew Zisserman.
\newblock Synthetic data for text localisation in natural images.
\newblock In {\em CVPR}, pages 2315--2324, 2016.

\bibitem{he2020momentum}
Kaiming He, Haoqi Fan, Yuxin Wu, Saining Xie, and Ross Girshick.
\newblock Momentum contrast for unsupervised visual representation learning.
\newblock In {\em Proceedings of the IEEE/CVF Conference on Computer Vision and
  Pattern Recognition}, pages 9729--9738, 2020.

\bibitem{he2019rethinking}
Kaiming He, Ross Girshick, and Piotr Doll{\'a}r.
\newblock Rethinking imagenet pre-training.
\newblock In {\em Proceedings of the IEEE/CVF International Conference on
  Computer Vision}, pages 4918--4927, 2019.

\bibitem{he2016deep}
Kaiming He, Xiangyu Zhang, Shaoqing Ren, and Jian Sun.
\newblock Deep residual learning for image recognition.
\newblock In {\em Proceedings of the IEEE conference on computer vision and
  pattern recognition}, pages 770--778, 2016.

\bibitem{he2021most}
Minghang He, Minghui Liao, Zhibo Yang, Humen Zhong, Jun Tang, Wenqing Cheng,
  Cong Yao, Yongpan Wang, and Xiang Bai.
\newblock Most: A multi-oriented scene text detector with localization
  refinement.
\newblock In {\em Proceedings of the IEEE/CVF Conference on Computer Vision and
  Pattern Recognition}, pages 8813--8822, 2021.

\bibitem{hendrycks2016gaussian}
Dan Hendrycks and Kevin Gimpel.
\newblock Gaussian error linear units (gelus).
\newblock {\em arXiv preprint arXiv:1606.08415}, 2016.

\bibitem{jia2021scaling}
Chao Jia, Yinfei Yang, Ye Xia, Yi-Ting Chen, Zarana Parekh, Hieu Pham, Quoc~V
  Le, Yunhsuan Sung, Zhen Li, and Tom Duerig.
\newblock Scaling up visual and vision-language representation learning with
  noisy text supervision.
\newblock {\em arXiv preprint arXiv:2102.05918}, 2021.

\bibitem{karatzas2015icdar}
Dimosthenis Karatzas, Lluis Gomez-Bigorda, Anguelos Nicolaou, Suman Ghosh,
  Andrew Bagdanov, Masakazu Iwamura, Jiri Matas, Lukas Neumann,
  Vijay~Ramaseshan Chandrasekhar, Shijian Lu, et~al.
\newblock Icdar 2015 competition on robust reading.
\newblock In {\em 2015 13th International Conference on Document Analysis and
  Recognition (ICDAR)}, pages 1156--1160. IEEE, 2015.

\bibitem{kim2021vilt}
Wonjae Kim, Bokyung Son, and Ildoo Kim.
\newblock Vilt: Vision-and-language transformer without convolution or region
  supervision.
\newblock {\em arXiv preprint arXiv:2102.03334}, 2021.

\bibitem{li2021align}
Junnan Li, Ramprasaath~R Selvaraju, Akhilesh~Deepak Gotmare, Shafiq Joty,
  Caiming Xiong, and Steven Hoi.
\newblock Align before fuse: Vision and language representation learning with
  momentum distillation.
\newblock {\em arXiv preprint arXiv:2107.07651}, 2021.

\bibitem{li2019visualbert}
Liunian~Harold Li, Mark Yatskar, Da Yin, Cho-Jui Hsieh, and Kai-Wei Chang.
\newblock Visualbert: A simple and performant baseline for vision and language.
\newblock {\em arXiv preprint arXiv:1908.03557}, 2019.

\bibitem{li2021trocr}
Minghao Li, Tengchao Lv, Lei Cui, Yijuan Lu, Dinei Florencio, Cha Zhang,
  Zhoujun Li, and Furu Wei.
\newblock Trocr: Transformer-based optical character recognition with
  pre-trained models.
\newblock {\em arXiv preprint arXiv:2109.10282}, 2021.

\bibitem{li2018shape}
Xiang Li, Wenhai Wang, Wenbo Hou, Ruo-Ze Liu, Tong Lu, and Jian Yang.
\newblock Shape robust text detection with progressive scale expansion network.
\newblock {\em arXiv preprint arXiv:1806.02559}, 2018.

\bibitem{liao2020mask}
Minghui Liao, Guan Pang, Jing Huang, Tal Hassner, and Xiang Bai.
\newblock Mask textspotter v3: Segmentation proposal network for robust scene
  text spotting.
\newblock In {\em Computer Vision--ECCV 2020: 16th European Conference,
  Glasgow, UK, August 23--28, 2020, Proceedings, Part XI 16}, pages 706--722.
  Springer, 2020.

\bibitem{liao2018textboxes++}
Minghui Liao, Baoguang Shi, and Xiang Bai.
\newblock Textboxes++: A single-shot oriented scene text detector.
\newblock {\em IEEE transactions on image processing}, 27(8):3676--3690, 2018.

\bibitem{liao2017textboxes}
Minghui Liao, Baoguang Shi, Xiang Bai, Xinggang Wang, and Wenyu Liu.
\newblock Textboxes: A fast text detector with a single deep neural network.
\newblock In {\em Thirty-first AAAI conference on artificial intelligence},
  2017.

\bibitem{liao2020real}
Minghui Liao, Zhaoyi Wan, Cong Yao, Kai Chen, and Xiang Bai.
\newblock Real-time scene text detection with differentiable binarization.
\newblock In {\em Proceedings of the AAAI Conference on Artificial
  Intelligence}, volume~34, pages 11474--11481, 2020.

\bibitem{2017Feature}
T.~Y. Lin, P. Dollar, R. Girshick, K. He, B. Hariharan, and S. Belongie.
\newblock Feature pyramid networks for object detection.
\newblock In {\em CVPR}, 2017.

\bibitem{liu2016ssd}
Wei Liu, Dragomir Anguelov, Dumitru Erhan, Christian Szegedy, Scott Reed,
  Cheng-Yang Fu, and Alexander~C Berg.
\newblock Ssd: Single shot multibox detector.
\newblock In {\em European conference on computer vision}, pages 21--37.
  Springer, 2016.

\bibitem{long2019rethinking}
Shangbang Long, Yushuo Guan, Bingxuan Wang, Kaigui Bian, and Cong Yao.
\newblock Rethinking irregular scene text recognition.
\newblock {\em arXiv preprint arXiv:1908.11834}, 2019.

\bibitem{long2021scene}
Shangbang Long, Xin He, and Cong Yao.
\newblock Scene text detection and recognition: The deep learning era.
\newblock {\em International Journal of Computer Vision}, 129(1):161--184,
  2021.

\bibitem{long2018textsnake}
Shangbang Long, Jiaqiang Ruan, Wenjie Zhang, Xin He, Wenhao Wu, and Cong Yao.
\newblock Textsnake: A flexible representation for detecting text of arbitrary
  shapes.
\newblock In {\em Proceedings of the European conference on computer vision
  (ECCV)}, pages 20--36, 2018.

\bibitem{long2020unrealtext}
Shangbang Long and Cong Yao.
\newblock Unrealtext: Synthesizing realistic scene text images from the unreal
  world.
\newblock {\em arXiv preprint arXiv:2003.10608}, 2020.

\bibitem{loshchilov2017decoupled}
Ilya Loshchilov and Frank Hutter.
\newblock Decoupled weight decay regularization.
\newblock {\em arXiv preprint arXiv:1711.05101}, 2017.

\bibitem{lyu2018mask}
Pengyuan Lyu, Minghui Liao, Cong Yao, Wenhao Wu, and Xiang Bai.
\newblock Mask textspotter: An end-to-end trainable neural network for spotting
  text with arbitrary shapes.
\newblock In {\em Proceedings of the European Conference on Computer Vision
  (ECCV)}, pages 67--83, 2018.

\bibitem{db2022github}
MhLiao.
\newblock Pytorch implementation of {DBNet}.
\newblock \url{https://github.com/MhLiao/DB}, 2022.

\bibitem{micikevicius2017mixed}
Paulius Micikevicius, Sharan Narang, Jonah Alben, Gregory Diamos, Erich Elsen,
  David Garcia, Boris Ginsburg, Michael Houston, Oleksii Kuchaiev, Ganesh
  Venkatesh, et~al.
\newblock Mixed precision training.
\newblock {\em arXiv preprint arXiv:1710.03740}, 2017.

\bibitem{nayef2017icdar2017}
Nibal Nayef, Fei Yin, Imen Bizid, Hyunsoo Choi, Yuan Feng, Dimosthenis
  Karatzas, Zhenbo Luo, Umapada Pal, Christophe Rigaud, Joseph Chazalon, et~al.
\newblock Icdar2017 robust reading challenge on multi-lingual scene text
  detection and script identification-rrc-mlt.
\newblock In {\em 2017 14th IAPR International Conference on Document Analysis
  and Recognition (ICDAR)}, volume~1, pages 1454--1459. IEEE, 2017.

\bibitem{oord2018representation}
Aaron van~den Oord, Yazhe Li, and Oriol Vinyals.
\newblock Representation learning with contrastive predictive coding.
\newblock {\em arXiv preprint arXiv:1807.03748}, 2018.

\bibitem{radford2021learning}
Alec Radford, Jong~Wook Kim, Chris Hallacy, Aditya Ramesh, Gabriel Goh,
  Sandhini Agarwal, Girish Sastry, Amanda Askell, Pamela Mishkin, Jack Clark,
  et~al.
\newblock Learning transferable visual models from natural language
  supervision.
\newblock {\em arXiv preprint arXiv:2103.00020}, 2021.

\bibitem{ranzato2007unsupervised}
Marc'Aurelio Ranzato, Fu~Jie Huang, Y-Lan Boureau, and Yann LeCun.
\newblock Unsupervised learning of invariant feature hierarchies with
  applications to object recognition.
\newblock In {\em 2007 IEEE conference on computer vision and pattern
  recognition}, pages 1--8. IEEE, 2007.

\bibitem{ren2015faster}
Shaoqing Ren, Kaiming He, Ross Girshick, and Jian Sun.
\newblock Faster r-cnn: Towards real-time object detection with region proposal
  networks.
\newblock {\em Advances in neural information processing systems}, 28:91--99,
  2015.

\bibitem{east2022github}
SakuraRiven.
\newblock Pytorch implementation of {EAST}.
\newblock \url{https://github.com/SakuraRiven/EAST}, 2022.

\bibitem{sennrich2015neural}
Rico Sennrich, Barry Haddow, and Alexandra Birch.
\newblock Neural machine translation of rare words with subword units.
\newblock {\em arXiv preprint arXiv:1508.07909}, 2015.

\bibitem{shi2017detecting}
Baoguang Shi, Xiang Bai, and Serge Belongie.
\newblock Detecting oriented text in natural images by linking segments.
\newblock In {\em Proceedings of the IEEE conference on computer vision and
  pattern recognition}, pages 2550--2558, 2017.

\bibitem{shrivastava2016training}
Abhinav Shrivastava, Abhinav Gupta, and Ross Girshick.
\newblock Training region-based object detectors with online hard example
  mining.
\newblock In {\em Proceedings of the IEEE conference on computer vision and
  pattern recognition}, pages 761--769, 2016.

\bibitem{singh2021textocr}
Amanpreet Singh, Guan Pang, Mandy Toh, Jing Huang, Wojciech Galuba, and Tal
  Hassner.
\newblock Textocr: Towards large-scale end-to-end reasoning for
  arbitrary-shaped scene text.
\newblock In {\em Proceedings of the IEEE/CVF Conference on Computer Vision and
  Pattern Recognition}, pages 8802--8812, 2021.

\bibitem{tang2019seglink++}
Jun Tang, Zhibo Yang, Yongpan Wang, Qi Zheng, Yongchao Xu, and Xiang Bai.
\newblock Seglink++: Detecting dense and arbitrary-shaped scene text by
  instance-aware component grouping.
\newblock {\em Pattern recognition}, 96:106954, 2019.

\bibitem{tian2019learning}
Zhuotao Tian, Michelle Shu, Pengyuan Lyu, Ruiyu Li, Chao Zhou, Xiaoyong Shen,
  and Jiaya Jia.
\newblock Learning shape-aware embedding for scene text detection.
\newblock In {\em Proceedings of the IEEE/CVF Conference on Computer Vision and
  Pattern Recognition}, pages 4234--4243, 2019.

\bibitem{wan2021self}
Qi Wan, Haoqin Ji, and Linlin Shen.
\newblock Self-attention based text knowledge mining for text detection.
\newblock In {\em CVPR}, pages 5983--5992, 2021.

\bibitem{pse2022github}
whai362.
\newblock Pytorch implementation of {PSENet}.
\newblock \url{https://github.com/whai362/PSENet}, 2022.

\bibitem{xu2019geometry}
Youjiang Xu, Jiaqi Duan, Zhanghui Kuang, Xiaoyu Yue, Hongbin Sun, Yue Guan, and
  Wayne Zhang.
\newblock Geometry normalization networks for accurate scene text detection.
\newblock In {\em Proceedings of the IEEE/CVF International Conference on
  Computer Vision}, pages 9137--9146, 2019.

\bibitem{xu2020layoutlm}
Yiheng Xu, Minghao Li, Lei Cui, Shaohan Huang, Furu Wei, and Ming Zhou.
\newblock Layoutlm: Pre-training of text and layout for document image
  understanding.
\newblock In {\em Proceedings of the 26th ACM SIGKDD International Conference
  on Knowledge Discovery \& Data Mining}, pages 1192--1200, 2020.

\bibitem{xu2019textfield}
Yongchao Xu, Yukang Wang, Wei Zhou, Yongpan Wang, Zhibo Yang, and Xiang Bai.
\newblock Textfield: Learning a deep direction field for irregular scene text
  detection.
\newblock {\em IEEE Transactions on Image Processing}, 28(11):5566--5579, 2019.

\bibitem{xu2020layoutlmv2}
Yang Xu, Yiheng Xu, Tengchao Lv, Lei Cui, Furu Wei, Guoxin Wang, Yijuan Lu,
  Dinei Florencio, Cha Zhang, Wanxiang Che, et~al.
\newblock Layoutlmv2: Multi-modal pre-training for visually-rich document
  understanding.
\newblock {\em arXiv preprint arXiv:2012.14740}, 2020.

\bibitem{yang2021tap}
Zhengyuan Yang, Yijuan Lu, Jianfeng Wang, Xi Yin, Dinei Florencio, Lijuan Wang,
  Cha Zhang, Lei Zhang, and Jiebo Luo.
\newblock Tap: Text-aware pre-training for text-vqa and text-caption.
\newblock In {\em Proceedings of the IEEE/CVF Conference on Computer Vision and
  Pattern Recognition}, pages 8751--8761, 2021.

\bibitem{yao2014unified}
Cong Yao, Xiang Bai, and Wenyu Liu.
\newblock A unified framework for multioriented text detection and recognition.
\newblock {\em IEEE Transactions on Image Processing}, 23(11):4737--4749, 2014.

\bibitem{yao2012detecting}
Cong Yao, Xiang Bai, Wenyu Liu, Yi Ma, and Zhuowen Tu.
\newblock Detecting texts of arbitrary orientations in natural images.
\newblock In {\em 2012 IEEE conference on computer vision and pattern
  recognition}, pages 1083--1090. IEEE, 2012.

\bibitem{yuliang2017detecting}
Liu Yuliang, Jin Lianwen, Zhang Shuaitao, and Zhang Sheng.
\newblock Detecting curve text in the wild: New dataset and new solution.
\newblock {\em arXiv preprint arXiv:1712.02170}, 2017.

\bibitem{zhou2017east}
Xinyu Zhou, Cong Yao, He Wen, Yuzhi Wang, Shuchang Zhou, Weiran He, and Jiajun
  Liang.
\newblock East: an efficient and accurate scene text detector.
\newblock In {\em Proceedings of the IEEE conference on Computer Vision and
  Pattern Recognition}, pages 5551--5560, 2017.

\end{thebibliography}
}

\end{document}